\documentclass[10pt,twocolumn,letterpaper]{article}

\usepackage{cvpr}              

\usepackage{graphicx}
\usepackage{amsmath}
\usepackage{amssymb}
\usepackage{booktabs}
\usepackage{float}
\usepackage{caption}
\usepackage{colortbl}
\usepackage{xcolor}
\usepackage{float}
\usepackage[export]{adjustbox}
\usepackage[symbol]{footmisc}

\definecolor{cvprblue}{rgb}{0.21,0.49,0.74}
\usepackage[pagebackref,breaklinks,colorlinks,citecolor=cvprblue]{hyperref}

\usepackage[capitalize]{cleveref}
\crefname{section}{Sec.}{Secs.}
\Crefname{section}{Section}{Sections}
\Crefname{table}{Table}{Tables}
\crefname{table}{Tab.}{Tabs.}

\newcommand{\cM}{\mathcal{M}}

\newcommand{\cV}{\mathcal{V}}

\newcommand{\cZ}{\mathcal{Z}}

\newcommand{\figref}[1]{Fig.~\ref{#1}}

\newcommand{\eqnref}[1]{Eq.~\eqref{#1}}

\makeatletter
\DeclareRobustCommand\onedot{\futurelet\@let@token\@onedot}
\def\@onedot{\ifx\@let@token.\else.\null\fi\xspace}
\def\eg{e.g\onedot} 
\def\ie{i.e\onedot} 
 
\def\etc{etc\onedot}

\def\etal{et~al\onedot}

\makeatother

\newcommand{\PAR}[1]{\vspace{0.1cm}\noindent{\bf #1} }

\def\cvprPaperID{11335} 
\def\confName{CVPR}
\def\confYear{2024}

\begin{document}
	\title{MVControl: Adding Conditional Control to Multi-view Diffusion for Controllable Text-to-3D Generation}
	\author{Zhiqi Li$^{1,2}$\qquad Yiming Chen$^{2,3}$ \qquad Lingzhe Zhao$^{2}$ \qquad Peidong Liu$^{2,\dag}$\vspace{0.1cm} \\
     $^{1}$Zhejiang University 
     \qquad 
     $^{2}$Westlake University 
     \qquad 
     $^{3}$Tongji University \\
		{\tt\small \{lizhiqi49, chenyiming, zhaolingzhe, liupeidong\}@westlake.edu.cn}
   }	
   \maketitle

   \let\thefootnote\relax\footnotetext{$^{\dag}$ Corresponding author.}

	
	\begin{abstract}
We introduce MVControl, a novel neural network architecture that enhances existing pre-trained multi-view 2D diffusion models by incorporating additional input conditions, \eg edge maps. Our approach enables the generation of controllable multi-view images and view-consistent 3D content. To achieve controllable multi-view image generation, we leverage MVDream as our base model, and train a new neural network module as additional plugin for end-to-end task-specific condition learning. 
To precisely control the shapes and views of generated images, we innovatively propose a new conditioning mechanism that predicts an embedding encapsulating the input spatial and view conditions, which is then injected to the network globally. Once MVControl is trained, score-distillation (SDS) loss based optimization can be performed to generate 3D content, in which process we propose to use a hybrid diffusion prior. The hybrid prior relies on a pre-trained Stable-Diffusion network and our trained MVControl for additional guidance. 
Extensive experiments demonstrate that our method achieves robust generalization and enables the controllable generation of high-quality 3D content.
Code available at \href{https://github.com/WU-CVGL/MVControl}{\textbf{\url{https://github.com/WU-CVGL/MVControl}}}.

\end{abstract}    
	\section{Introduction}
\label{sec:intro}

\begin{figure*}\label{fig_teaser}
	\centering
	\includegraphics[width=0.93\linewidth]{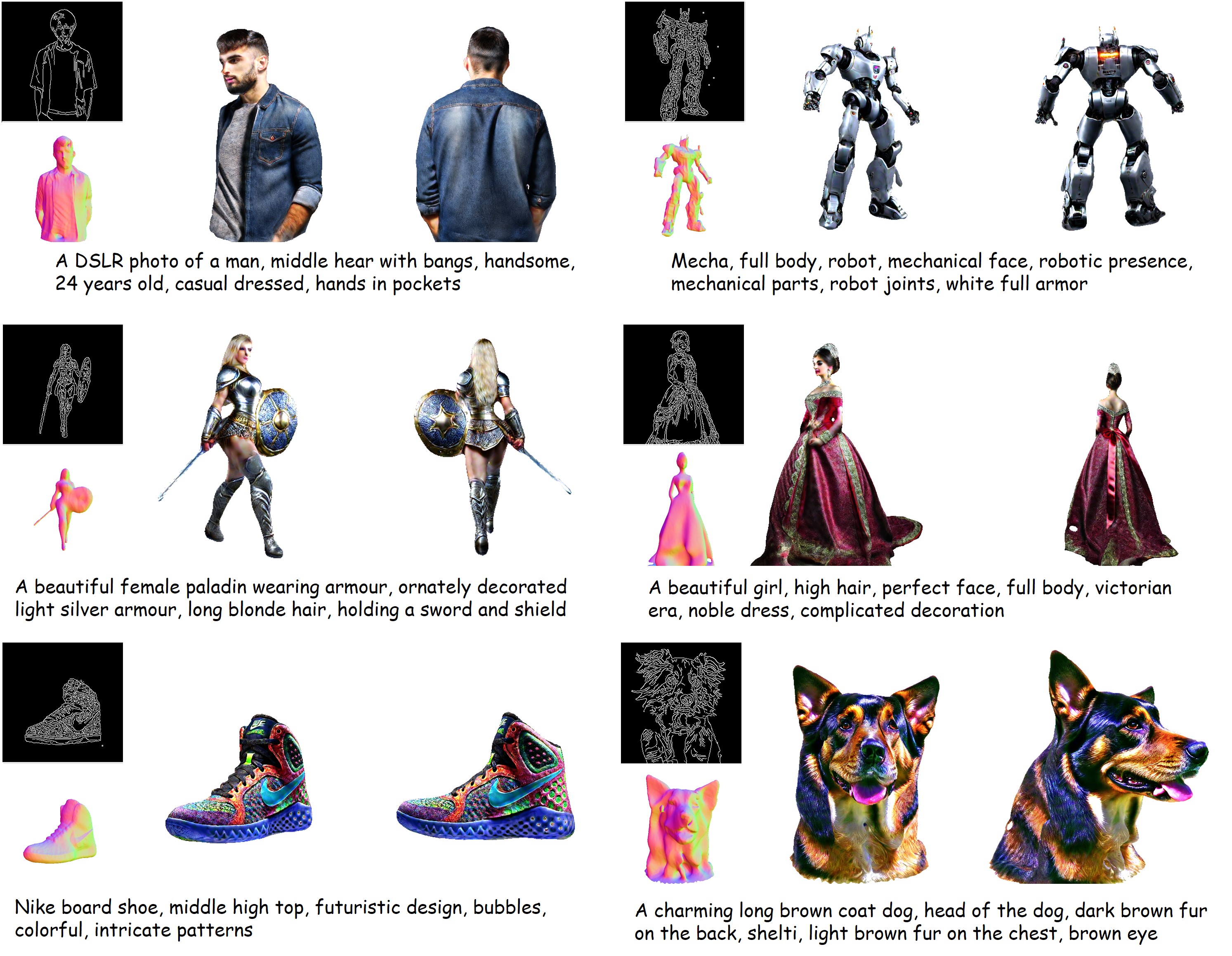}
	\vspace{-0.5em}
	\captionof{figure}{\textbf{MVControl:} Given an input text prompt and an edge map, our method is able to generate high-fidelity controllable multi-view images and view-consistent 3D content.}
 \vspace{-1em}
\end{figure*}

Remarkable progress has been made in the field of 2D image generation recently. High fidelity images can be easily generated via input text prompts \cite{rombach2022stablediffusion}. Due to the scarcity of 3D training data, the success in text-to-image generation is hardly copied to the text-to-3D domain. 
Instead of training a large text-to-3D generative model from scratch with large amounts of 3D data, due to the nice properties of diffusion models \cite{ho2020denoising, song2020score} and differentiable 3D representations \cite{mildenhall2021nerf, wang2021neus, shen2021dmtet, kerbl20233dgaussian}, recent score distillation optimization (SDS) \cite{poole2022dreamfusion} based methods \cite{poole2022dreamfusion, lin2023magic3d, tang2023make, chen2023fantasia3d, metzer2023latent, wang2023prolificdreamer, tang2023dreamgaussian}, attempt to distill 3D knowledge from a pre-trained large text-to-image generative model \cite{rombach2022stablediffusion} and have achieved impressive results. The representative work is DreamFusion \cite{poole2022dreamfusion}, which starts a new paradigm for 3D asset generation. 


Following the 2D-to-3D distillation methodology, the techniques are rapidly evolving over the past year. Many works have been proposed to further improve the generation quality by applying multiple optimization stages \cite{lin2023magic3d, chen2023fantasia3d}, optimizing the diffusion prior with the 3D representation simultaneously \cite{wang2023prolificdreamer, sun2023dreamcraft3d}, deriving more precise formulation of score distillation algorithm \cite{katzir2023nfsd, yu2023csd}, or enhancing the details of whole pipeline \cite{huang2023dreamtime, armandpour2023prepneg, zhu2023hifa}. 
Although the above mentioned efforts can get access to delicate texture, the view consistency of generated 3D content is hard to be achieved, since the 2D diffusion prior is not view-dependent. Hence, there is a series of works hammering at introducing multi-view knowledge to the pre-trained diffusion models \cite{liu2023zero123, shi2023mvdream, liu2023syncdreamer, li2023sweetdreamer, shi2023zero123++, long2023wonder3d}. 
Although they can deliver impressive text controlled multiview images and 3D assets, they still cannot achieve fine-grained control over the generated content via an edge map for example, as its counterpart in text-to-image generation, \ie ControlNet \cite{zhang2023controlnet}. 
%
In this work, we therefore propose MVControl, a multi-view version of ControlNet, to enable controllable text-to-multi-view image  generation. Once MVControl is trained, we can exploit it to the score distillation optimization pipeline, so as to achieve controllable text-to-3D content generation via an input condition image, \eg edge map.

Inspired by 2D ControlNet \cite{zhang2023controlnet}, which works as a plug-in module of Stable-Diffusion \cite{rombach2022stablediffusion}, we choose a recently released multi-view diffusion network, MVDream \cite{shi2023mvdream}, as our base model. A control network is then designed to interact with the base model to achieve controllable text-to-multi-view image generation. Similarly to \cite{zhang2023controlnet}, the weights of MVDream is all frozen and we only train the control network.
While MVDream is trained with camera poses defined in the absolute world coordinate system, we experimentally find that the 
relative pose condition with respect to the condition image is more proper for controllable text-to-multi-view generation. However, it conflicts with the definition of the pretrained MVDream network. Furthermore, since the conditioning mechanism of 2D ControlNet is designed for single image generation and does not consider the multi-view scenario, view-consistency cannot be easily achieved by directly adopting its control network to interact with the base model.
To overcome these issues, we design a simple but effective novel conditioning strategy based on the original ControlNet architecture to achieve controllable text-to-multi-view generation. MVControl is jointly trained on a subset of the large-scale 2D dataset LAION \cite{schuhmann2022laion} and 3D dataset Objaverse \cite{deitke2023objaverse} as what \cite{shi2023mvdream} does. We only explore to use the edge map as conditional input in this work. However, our network has no restriction to use other type of input conditions, \eg depth map, sketch image \etc. 
Once MVControl is trained, we can exploit it to provide 3D priors for controllable text-to-3D asset generation. In particular, we employ a hybrid diffusion prior relying on pretrained Stable-Diffusion model and MVControl network. The generation is conducted in a coarse-to-fine manner. After we get a good geometry after the coarse stage, we fix it and only optimize the texture during the fine stage. 
Our extensive experiments demonstrate that our proposed method can generate fine-grain controlled high-fidelity multi-view images and 3D content via an input condition image as well as textual description.
 
In summary, our main contributions are as follows.
\begin{itemize}
	\itemsep0em
    \item We propose a novel network architecture to achieve fine-grain controlled text-to-multi-view image generation; 
    \item Once our network is trained, it can be exploited to serve as a part of hybrid diffusion prior for controllable text-to-3D content generation vis SDS optimization;
    \item Extensive experimental results demonstrate that our method is able to deliver high-fidelity multi-view image and 3D asset, which can be fine-grain controlled by an input condition image and text prompt;
    \item Besides being used to generate 3D asset via SDS optimization, we believe our MVControl network could benefit the general 3D vision/graphic community for broad application scenarios.
\end{itemize}

	\section{Related Work}
\label{sec:related}

We review the related works in 3D generation and classify them into three categories: diffusion-based novel view synthesis, 3D generative models and text-to-3D generation, which are the most related to our method.

\PAR{Diffusion-based novel view synthesis.} The success of text-to-image generation via large diffusion models inspires the development of pose-guided novel view image synthesis. Commonly adopted approach is to condition on a diffusion model by an additional input image and target pose \cite{liu2023zero123,watson2023novelview,tseng2023novelview,xiang20233daware}. Different from those methods, Chan \etal recently proposes to learn 3D scene representation from a single or multiple input images and then exploit a diffusion model for target novel view image synthesis \cite{chan20233diffusion}. Instead of generate a single target view image, MVDiffusion \cite{tang2023mvdiffusion} proposes to generate multi-view consistent images in one feed-forward pass. They build upon a pre-trained diffusion model to have better generalization capability. MVDream \cite{shi2023mvdream} also proposes to generate consistent multi-view images from a text prompt recently, by fine-tuning a pre-trained diffusion model with a 3D dataset. They then exploit the trained model to serve as a 3D prior to optimize the 3D representation via Score Distillation Sampling. While prior work can generate impressive novel/multi-view consistent images, fine-grained control over the generated text-to-multi-view images is still difficult to achieve, as what ControlNet \cite{zhang2023controlnet} has achieved for text-to-image generation. Therefore, we propose a multi-view ControlNet (\ie MVControl) in this work to further advance diffusion-based multi-view image generation. 

\PAR{3D generative models.} Current 3D generative models usually exploit existing 3D datasets to train generative models with different 3D representations. Commonly used 3D representations are volumetric representation \cite{wu2016learning,brock20163dgan,gadelha20173d,li2019gan}, triangular mesh \cite{ben2018multi,tan2018cvpr,gao2019tog,pavllo2021iccv,youwang2022eccv}, point cloud \cite{pumarola2020cvpr, nichol2022point, achlioptas2018learning, shu20193d, yang2019pointflow} as well as the recent implicit neural representation \cite{park2019cvpr,mescheder2019cvpr,chen2019cvpr,schwarz2022neurips,chan2022eg3d,wang2023rodin}. 
Various generative modeling techniques have also been explored to 3D data as their success in 2D image synthesis, which range from variational auto-encoder \cite{balashova2018structure,wu2019ToG,tan2018cvpr,gao2019tog}, generative adversarial network \cite{wu2016learning,li2019gan,phuoc2020neurips,pavllo2021iccv,achlioptas2018learning,chan2022eg3d}, flow-based method \cite{aliakbarian2022cvpr,yang2019iccv,pumarola2020cvpr,klokov2020eccv}, and the recent popular diffusion based method \cite{luo2021cvpr,zhou2021iccv,zeng2022neurips,hui2022neural,muller2023diffrf,chou2023diffsdf}. Different from image generative modeling, which has large amount of training images, those 3D generative methods usually lack sufficient 3D assets for training. They are usually limited to category-specific dataset, \eg shapeNet \cite{shapenet2015}. Although Objaverse \cite{deitke2023objaverse} released a million-scale 3D asset dataset recently, its size is still infancy compared to the 2D training data \cite{schuhmann2022laion} used by modern generative models for image synthesis. 
Due to the lack of large amount of training data, they usually cannot generate arbitrary type of objects to satisfy the requirements of end consumers. Instead of relying on large amount of 3D data as those methods, we propose to exploit a pre-trained large image generative model to distill 3D knowledge for controlled text-to-3D generation. 

\PAR{Text-to-3D generation.} Due to the scarcity of 3D data, researchers attempt to distill knowledge for 3D generation from pre-trained large image models. The initial attempt was to exploit a pre-trained CLIP \cite{radford2021learning} model to align the input text prompt and rendered images for the supervision of 3D object generation \cite{sanghi2022clipforge, jain2022zero, mohammad2022clip}. However, the generated 3D objects usually tend to be less realistic due to that CLIP \cite{radford2021learning} can only offer high-level semantic guidance.  
With the advancement of large text-to-image diffusion models \cite{saharia2022photorealistic}, DreamFusion \cite{poole2022dreamfusion} demonstrates the potential to generate more realistic 3D objects via knowledge distillation. Follow-up works continue to push the performance to generate photo-realistic 3D objects that closely match the provided text prompts \cite{lin2023magic3d,wang2023SJC,chen2023fantasia3d,wang2023prolificdreamer,li2023sweetdreamer,tang2023dreamgaussian,zhu2023hifa,huang2023dreamtime,raj2023dreambooth3d}. The main insights of those methods are usually to develop more advanced score distillation loss or better optimization strategy \etc. to further improve the generation quality.
Although those methods generate high-fidelity 3D shapes via text description, fine-grained control on the text-to-3D shape generation is still lacking. We therefore propose to exploit our pre-trained MVControl network to provide a 3D prior for controllable text-to-3D generation.

	\section{Method}
\label{sec:method}

\begin{figure*}[!htbp]
    \centering
    \begin{minipage}[b]{0.45\textwidth}
        \centering
        \begin{subfigure}[b]{0.95\textwidth}
            \includegraphics[width=\textwidth]{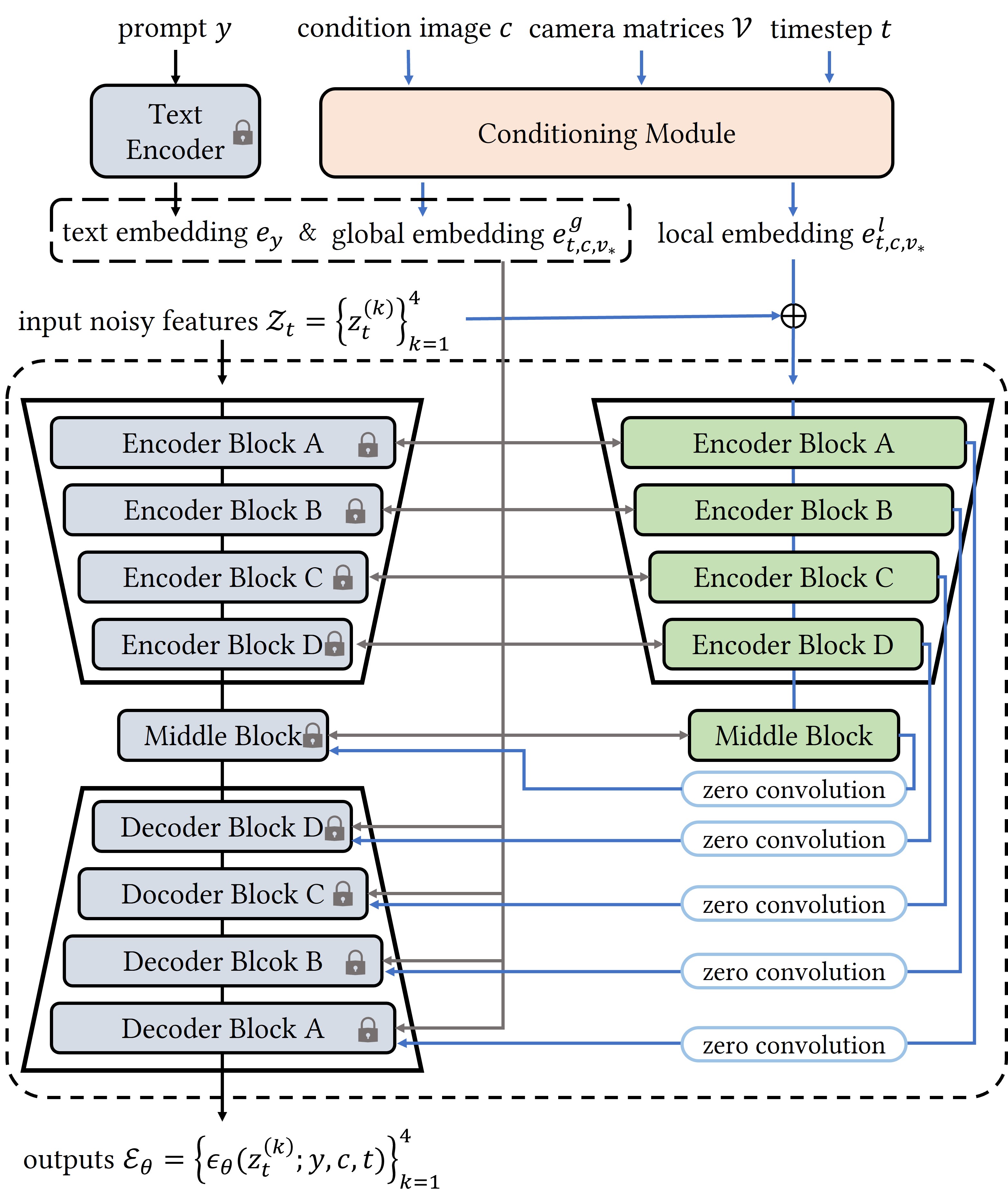}
            \caption{Framework of MVControl}
            \label{fig_framework} 
        \end{subfigure}
    \end{minipage}\hspace{5mm}%
    \begin{minipage}[b]{0.5\textwidth}
        \centering
        \begin{minipage}[b]{\textwidth}
            \centering
            \begin{subfigure}[b]{0.95\textwidth}
                \includegraphics[width=\textwidth]{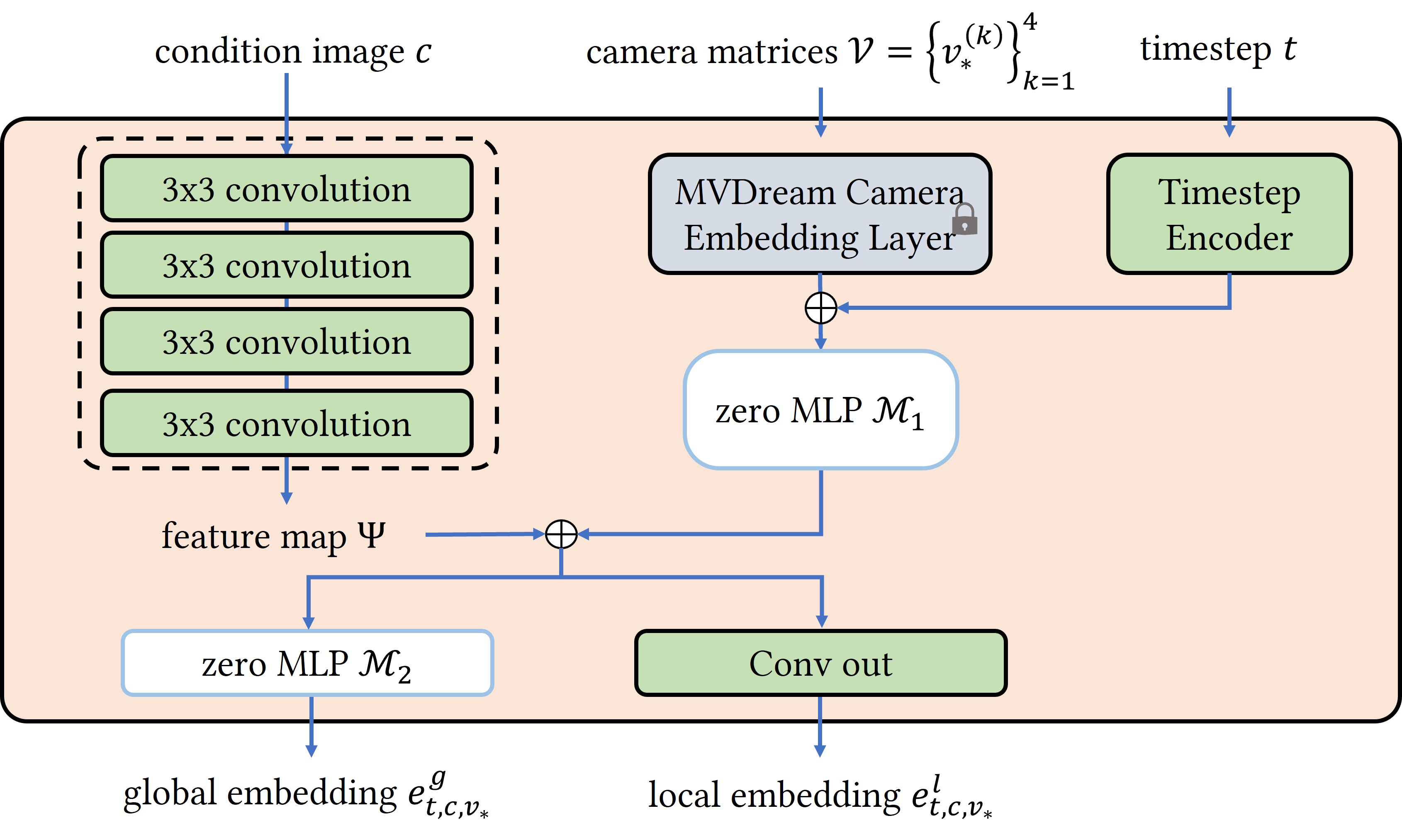}
                \caption{Architecture of conditioning module}
                \label{fig_cond_module} 
            \end{subfigure}
        \end{minipage}\vspace{5mm}
        \begin{minipage}[b]{\textwidth}
            \centering
            \begin{subfigure}[b]{0.95\textwidth}
                \includegraphics[width=\textwidth]{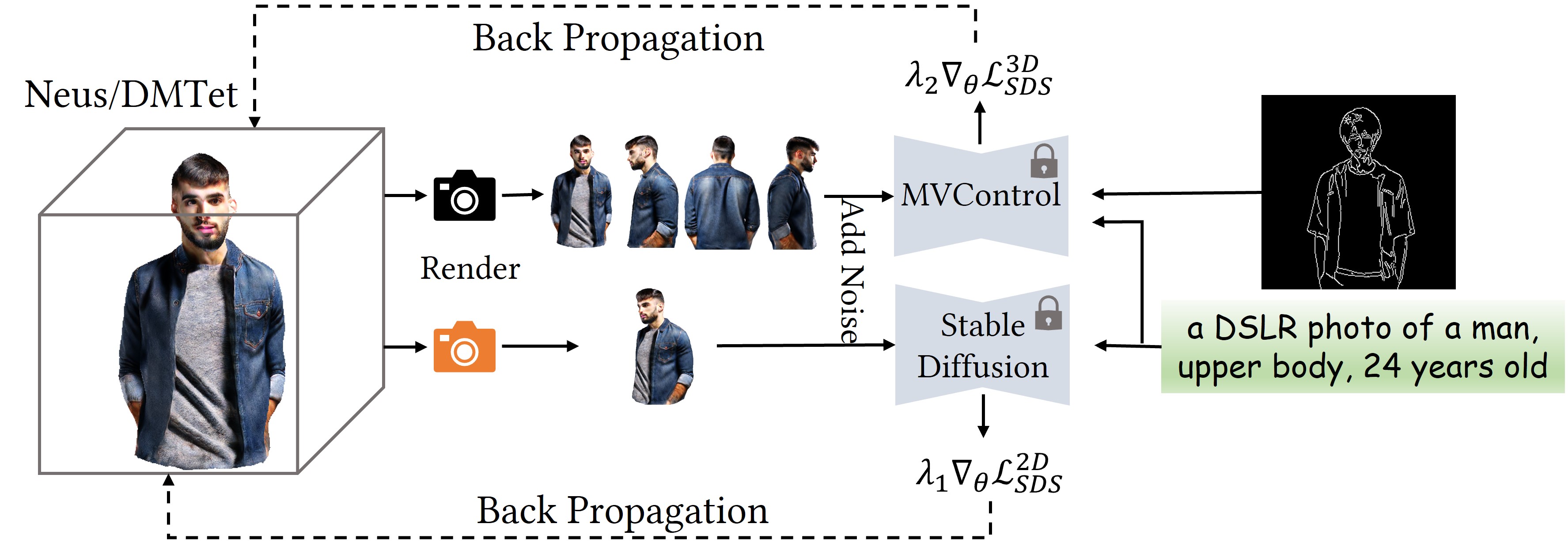}
                \caption{Controllable 3D generation with hybrid diffusion prior}
                \label{fig_3d_gen}
            \end{subfigure}
        \end{minipage}
    \end{minipage}
    \caption{{\bf{Overview of proposed method.}} (a) MVControl consists of a frozen multi-view diffusion model and a trainable MVControl. (b) Our model takes care of all input conditions to control the generation process both locally and globally through a conditioning module. (c) Once MVControl is trained, we can exploit it to serve a hybrid diffusion prior for controllable text-to-3D content generation via SDS optimization procedure.}
    \label{fig_mvcontrol}
    \vspace{-0.2em}
\end{figure*}

We first review relevant methods, including diffusion model \cite{ho2020denoising, song2020score},  MVDream \cite{shi2023mvdream}, ControlNet \cite{zhang2023controlnet} and score distillation sampling \cite{poole2022dreamfusion} in Section \ref{subsec:preliminary}. Then, we analyze the strategy of introducing additional spatial conditioning to MVDream by training a multi-view ControlNet in Section \ref{subsec:MVControl}. Finally in Section \ref{subsec:3dgen}, based on the trained multi-view ControlNet, we propose the realization of controllable 3D content generation using SDS loss with hybrid diffusion priors as guidance.

\subsection{Preliminary}
\label{subsec:preliminary}

\PAR{Diffusion model.} Diffusion model predicts the score function $\nabla_{\mathbf{x_t}}\log p_{data}(\mathbf{x_t})$ in the data space under different noise level $t\sim \mathcal{U}(0, T)$, so as to guide the sampling process to progressively denoise a pure noise $x_T\sim \mathcal{N}(\mathbf{0}, \mathbf{I})$ to a clean data $x_0$. To learn the denoising score, noises at different scales are added to $x_0$ with pre-defined noise schedule according to:
\begin{equation}\label{eq:1}
    x_t=\sqrt{\bar{\alpha}_t}x_0 + \sqrt{1-\bar{\alpha}_t}\epsilon, 
\end{equation}
where $\alpha_t\in(0,1)$, $\bar{\alpha}_t=\prod_{s=1}^{t}\alpha_{s}$ and $\epsilon \sim \mathcal{N}(\mathbf{0}, \mathbf{I})$. The diffusion model parameterized by $\phi$ can then be trained by minimizing the noise reconstruction loss:
\begin{equation}
    \mathcal{L}_{\textrm{diffusion}}=\mathbb{E}_{t,\epsilon}\lbrack\Vert \epsilon_{\phi}(x_t, t)-\epsilon \Vert _2 ^2\rbrack. \label{eq:2}
\end{equation}
Once the model is trained, it can be iteratively applied to denoise a sampled noise $x_T\sim \mathcal{N}(\mathbf{0}, \mathbf{I})$ to obtain a clean data. The readers can refer to \cite{ho2020denoising, song2020score} for more detailed illustration about diffusion models.

\PAR{MVDream.} MVDream \cite{shi2023mvdream} is a recently proposed text-to-multi-view diffusion model. It is fine-tuned based on a large-scale pre-trained image diffusion model, \ie Stable-Diffusion (SD) \cite{rombach2022stablediffusion}, via text labled multi-view image data. While SD generates one image for a sampling process, MVDream generates four images simultaneously. The four images corresponding to four consistent views of an object, and are 90 degrees apart from each other in the longitude direction with the same elevation. To ensure the interaction of features among different views for view-consistent generation, MVDream replaces self attention layers of SD with cross-view attentions, which concatenate the patches of all views before attention computation. To enable view conditioning, they further exploit a MLP (Multi-layer Perceptron) network to transform camera extrinsic parameters into an embedding vector, which are then concatenated with the timestep embedding to be injected to the whole model.

\PAR{ControlNet.} ControlNet \cite{zhang2023controlnet} enables pretrained large diffusion models to support additional input conditions (\eg canny edges, sketches, depth maps, \etc.) beside the text prompts. It is constructed by directly copying the structure and weights of SD's encoder blocks and mid block, and adding zero convolution layers to connect it with the pretrained SD. 
With those connections, the feature map computed by each inner layer of ControlNet can then be injected to its corresponding symmetric layer in SD's UNet decoder, so as to control the sampling process of SD once it is trained.

\PAR{Score distillation sampling.} Score distillation sampling (SDS) \cite{poole2022dreamfusion, lin2023magic3d} leverages pretrained text-to-image diffusion model as prior to guide text-conditioned 3D asset generation. In particular, given a pre-trained diffusion model $\epsilon_{\phi}$, SDS optimizes the parameters $\theta$ of a differentiable 3D representation, \eg neural radiance field, using the gradient of the loss $\mathcal{L}_{SDS}$ with respect to $\theta$:
\begin{equation}
    \nabla_{\theta}\mathcal{L}_{\textrm{SDS}}(\phi, \mathbf{x})=\mathbb{E}_{t,\epsilon}\lbrack w(t)(\hat{\epsilon}_\phi-\epsilon)\frac{\partial{z_t}}{\partial{\theta}}\rbrack, \label{eq:3}
\end{equation}
where $\mathbf{x}=g(\theta, c)$ is an image rendered by $g$ under a camera pose $c$, $w(t)$ is a weighting function depending on the timestep $t$ and $z_t$ is the noisy image input to diffusion model by adding Gaussian noise $\epsilon$ to $\mathbf{x}$ corresponding to the $t$-th timestep according to \eqnref{eq:1}. The main insight is to enforce the rendered image of the learnable 3D representation to satisfy the distribution of the pretrained diffusion model. In practice, the values of timestep $t$ and Gaussian noise $\epsilon$ are randomly sampled at every optimization step. 

\subsection{Multi-view ControlNet}\label{subsec:MVControl}
Inspired by ControlNet in controlled text-to-image generation and recently released text-to-multi-view image diffusion model (\eg MVDream), we aim to design a multi-view version of ControlNet (\ie MVControl) to achieve controlled text-to-multi-view generation. As shown in \figref{fig_framework}, we follow similar architecture style as ControlNet, \ie a locked pre-trained MVDream and a trainable control network. The main insight is to preserve the learned prior knowledge of MVDream, while train the control network to learn the inductive bias with small amount of data. The control network consists of a conditioning module and a copy of the encoder network of MVDream. Our main contribution lies at the conditioning module and we will detail it as follows. 

The conditioning module (\figref{fig_cond_module}) receives the condition image $c$, four camera matrices $\mathcal{V}_*\in\mathbb{R}^{4\times4\times4}$ and timestep $t$ as input, and outputs four local control embeddings $e^l_{t, c, v_*}$ and global control embeddings $e^g_{t, c, v_*}$. The local embedding is then added with the input noisy latent features $\cZ_t\in\mathbb{R}^{4\times C\times H\times W}$ as the input to the control network, and the global embedding $e^g_{t, c, v_*}$ is injected to each layer of MVDream and MVControl to globally control generation.

The condition image $c$ (\ie edge map, depth map \etc) is processed by four convolution layers to obtain a feature map $\Psi$. Instead of using the absolute camera pose matrices embedding of MVDream, we move the embedding into the conditioning module. To make the network better understand the spatial relationship among different views, the relative camera poses with respect to the condition image are used for the camera matrices $\cV_*$. The experimental results also validate the effectiveness of the design. The camera matrices embedding is combined with the timestep embedding, and is then mapped to have the same dimension as the feature map $\Psi$ by a zero-initialized module $\cM_1$. The sum of these two parts is projected to the local embedding $e^l_{t, c, v_*}$ through a convolution layer.

While MVDream is pretrained with absolute camera poses, the conditioning module exploit relative poses as input. We experimentally find that the network hardly converges due to the mismatch of both coordinate frames. We therefore exploit an additional network $\cM_2$ to learn the transformation and output a global embedding $e^g_{t, c, v_*}$ to replace the original camera matrix embedding of MVDream and add on timestep embeddings of both MVDream and MVControl part so that inject semantical and view-dependent features globally.

\subsection{Controllable 3D Content Generation}\label{subsec:3dgen}
Once MVControl is trained, it can be utilized for controllable text-to-3D content generation via SDS optimization \cite{poole2022dreamfusion}. We adopt a hybrid diffusion prior from both Stable-Diffusion and MVControl, to better guide the 3D generation. MVControl provides a strong consistent geometry guidance over four canonical views of the optimizing 3D object, while Stable-Diffusion provides fine geometry and texture sculpting at the other randomly sampled views. 
As is shown in \figref{fig_3d_gen}, the hybrid SDS gradient can be calculated as:
\begin{equation}
    \nabla_\theta\mathcal{L}_{SDS}^{hybrid}=\lambda_1 \nabla_\theta\mathcal{L}_{SDS}^{2D} + \lambda_2 \nabla_\theta\mathcal{L}_{SDS}^{3D}, \label{eq:hybrid_sds}
\end{equation}
where $\lambda_1$ and $\lambda_2$ are the strength of 2D and 3D prior respectively.
The optimization procedure consists of two stages: a coarse stage for initial model generation and a fine stage for texture refinement. 

During the coarse stage, we exploit a coarse neural surface, \ie NeuS \cite{wang2021neus}, to represent the 3D asset. SDS loss is then computed to optimize the neural 3D representation. To encourage the smoothness of the surface, we also exploit eikonal loss \cite{wang2021neus} to regularize the training process. To obtain a high-fidelity 3D asset, we extract the coarse neural surface to a hybrid mesh representation via DMTet \cite{shen2021dmtet}. The texture is further refined by SDS loss with fixed geometry. We use the conventional SDS loss \cite{poole2022dreamfusion} for coarse stage generation, and we use the recently proposed noise-free score distillation \cite{katzir2023nfsd} for our fine stage, which delivers similar performance with conventional SDS but can use normal scale for classifier-free guidance (CFG) \cite{ho2022cfg}.

	\section{Experiments} \label{sec:exp}

\begin{figure*}
	\centering
	\includegraphics[width=0.99\linewidth]{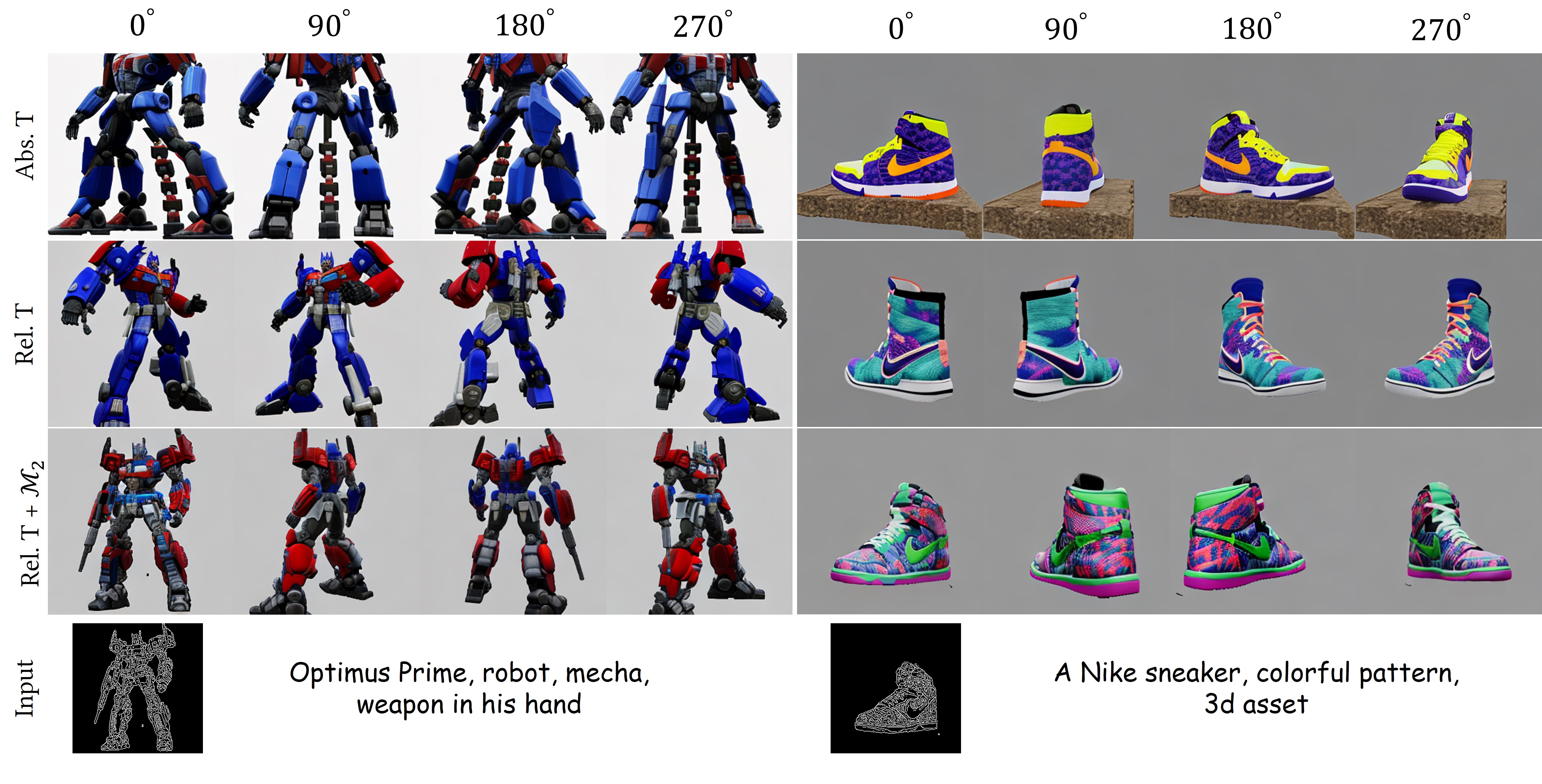}
	\vspace{-0.5em}
	\captionof{figure}{\textbf{The necessity on the design of camera pose condition.} It demonstrates that only the complete conditioning module can properly control the generation of the posed images, which is defined relative to the input condition edge image. \textbf{Abs. T} denotes the conditioning module does not accept any pose condition as input, the whole network relies on the absolute pose condition of MVDream for pose control; \textbf{Rel. T} denotes the MLP network $\cM_2$ is removed from the conditioning module, which is used to bridge the relative pose condition and the base model, which is pretrained with absolute pose condition; \textbf{Rel. T + $\cM_2$} denotes the complete module.}
        \label{fig_ablation_pose}
	\vspace{-1.0em}
\end{figure*}

\subsection{Implementation Details}
\label{subsec:impl}

\PAR{Data Preparation.}\label{subsubsec:impl_2d}
We use the multi-view renderings of the public large 3D dataset, Objaverse \cite{deitke2023objaverse} to train our MVControl. Firstly we clean the dataset by excluding all samples with CLIP-score lower than 22 based on the labeling of \cite{sun2023unig3d} and finally we have about 400k samples left. Instead of using the name and tags of the 3D assets, we refer to the captions of \cite{luo2023cap3d} as text descriptions of our kept objects. For each object, we first normalize its scene bounding box to unit cube at the world center, and then randomly sample camera distance between [1.4, 1.6], fov between [40, 60] and elevation between [0, 30]. Finally, we randomly sample from 32 uniformly distributed azimuths as starting point to sample 4 orthogonal views each time.

\PAR{Training details of MVControl network.}
We exploit the weights of pretrained MVDream and ControlNet to initialize our network. All the connections between the locked and trainable networks are initialized with zero. Our network is then trained with both 2D and 3D datasets. In particular, We sample images from the AES v2 subset of LAION \cite{schuhmann2022laion} with a 30\% probability for training, such that the network will not lose its learned 2D image priors. We then sample from our prepared 3D/mult-view image dataset with a 70\% probability to learn the 3D knowledge. We exploit the Canny edge map of a sampled image as the conditioning image for training. Other options, \eg depth image, sketch \etc, can also be exploited without any modification of the method. 

The training images have a resolution at 256x256 pixels, and batch size is chosen as 2560 images. The model is fine-tuned for 50,000 steps under a conservative learning rate, $4e\times10^{-5}$, on 8 Nvidia Tesla A100 GPUs with AdamW optimizer \cite{kingma2014adam}. Following \cite{zhang2023controlnet}, we also drop the text prompt of one sample as empty with 50\% chance for classifier-free training, such that the model can be trained to better understand the semantics of input condition images.

\PAR{3D Content Generation.}\label{subsubsec:impl_3d}
We choose Stable-Diffusion-v2.1-base \cite{rombach2022stablediffusion} as the 2D part of our hybrid diffusion prior. In the coarse stage, we use the NeuS 3D representation and Instant-NGP \cite{muller2022instant} as its implementation for training efficiency. The neural surface is optimized for 8,000 steps with AdamW optimizer. Its rendering resolution is increased from $64\times64$ to $256\times256$ after 5,000 steps. We also employ timestep annealing strategy. The timestep sampling range is gradually decreased from $(0.98,0.98)$ to $(0.5, 0.02)$ over the optimization process. The CFG scale for 2D and 3D part of our hybrid diffusion prior is empirically chosen as 10 and 50 respectively. For the computation of the SDS loss of MVControl, we use the $x_0$-reconstruction formulation proposed in \cite{shi2023mvdream} to use CFG rescale trick \cite{lin2023common}, and the rescale factor is set as 0.5.
In the fine stage, we extract the neural surface to 128 grid DMTet and the rendered image resolution is set to be $512\times512$ pixels. The CFG scale for 2D and 3D part is 7.5 and 10 respectively. For both stages, we set the strength of 2D diffusion guidance as 1, and that of MVControl guidance as 0.1 or 0.2 empirically. 

\begin{figure*}
	\centering
	\includegraphics[width=0.99\linewidth]{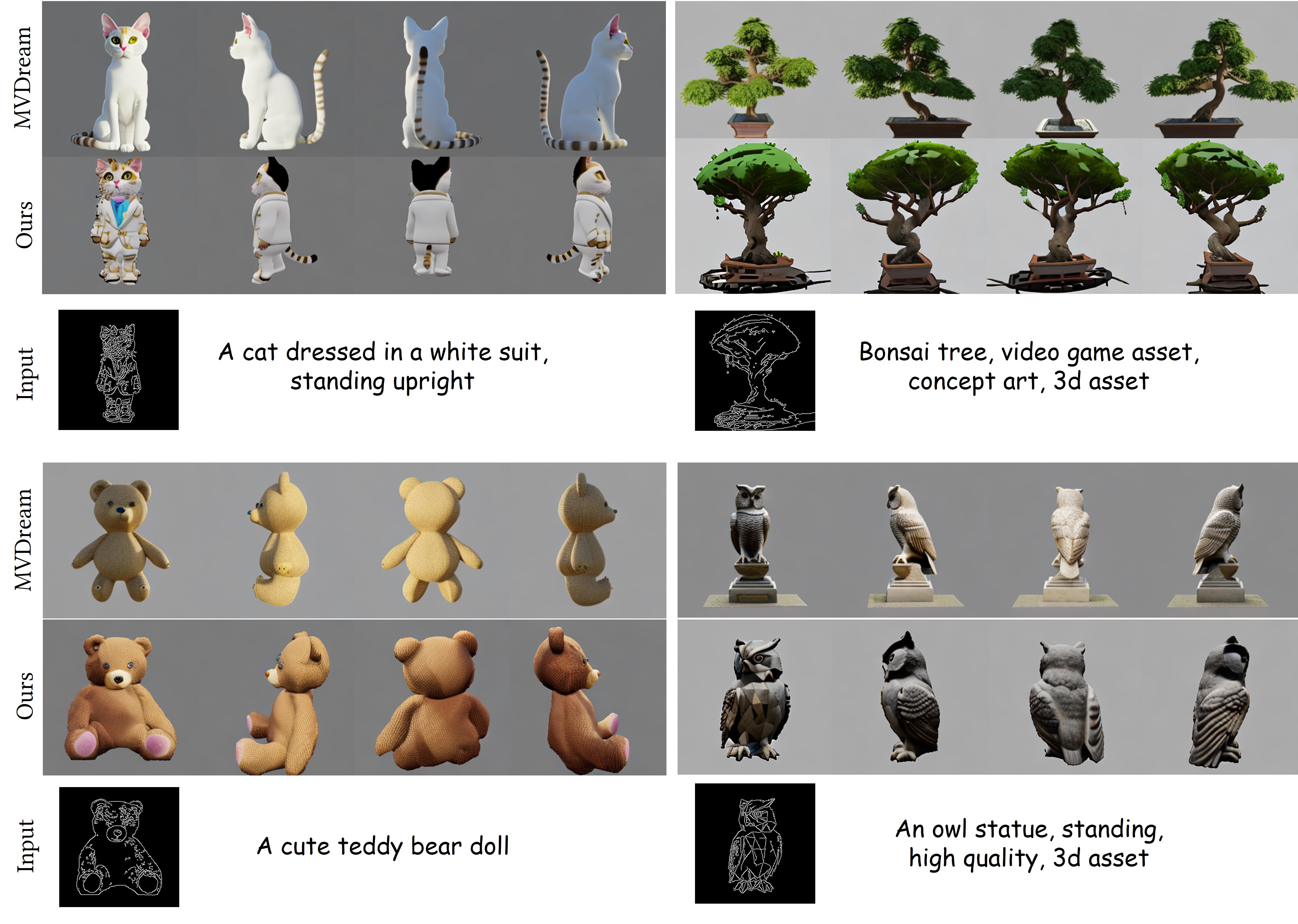}
	\captionof{figure}{\textbf{Controllable multi-view image generation.} It demonstrates that our method is able to generate controllable high-fidelity multi-view images, satisfying both the input condition image and text prompt.}
        \label{fig_2d_exp1}
	\vspace{-1em}
\end{figure*}


\subsection{Ablation Studies}
\PAR{The necessity on the design of camera pose condition.} We train our network under three different settings: 1) we exploit the absolute pose condition (\ie Abs. T) of MVDream \cite{shi2023mvdream}, and only have the local embedding of the input condition image as the output from the conditioning module; 2) we remove the zero MLP $\cM_2$ (\ie Rel. T), which is used to bridge the relative pose embedding of the conditioning module and that of MVDream base model; 3) complete conditioning module (\ie Rel. T+$\cM_2$). The experimental results are shown in \figref{fig_ablation_pose}, it demonstrates that only the complete conditioning module can have good control over the pose of the generated images. The pose is defined as relative pose with respect to the input condition image. 

\begin{figure*}
	\centering
	\includegraphics[width=0.99\linewidth]{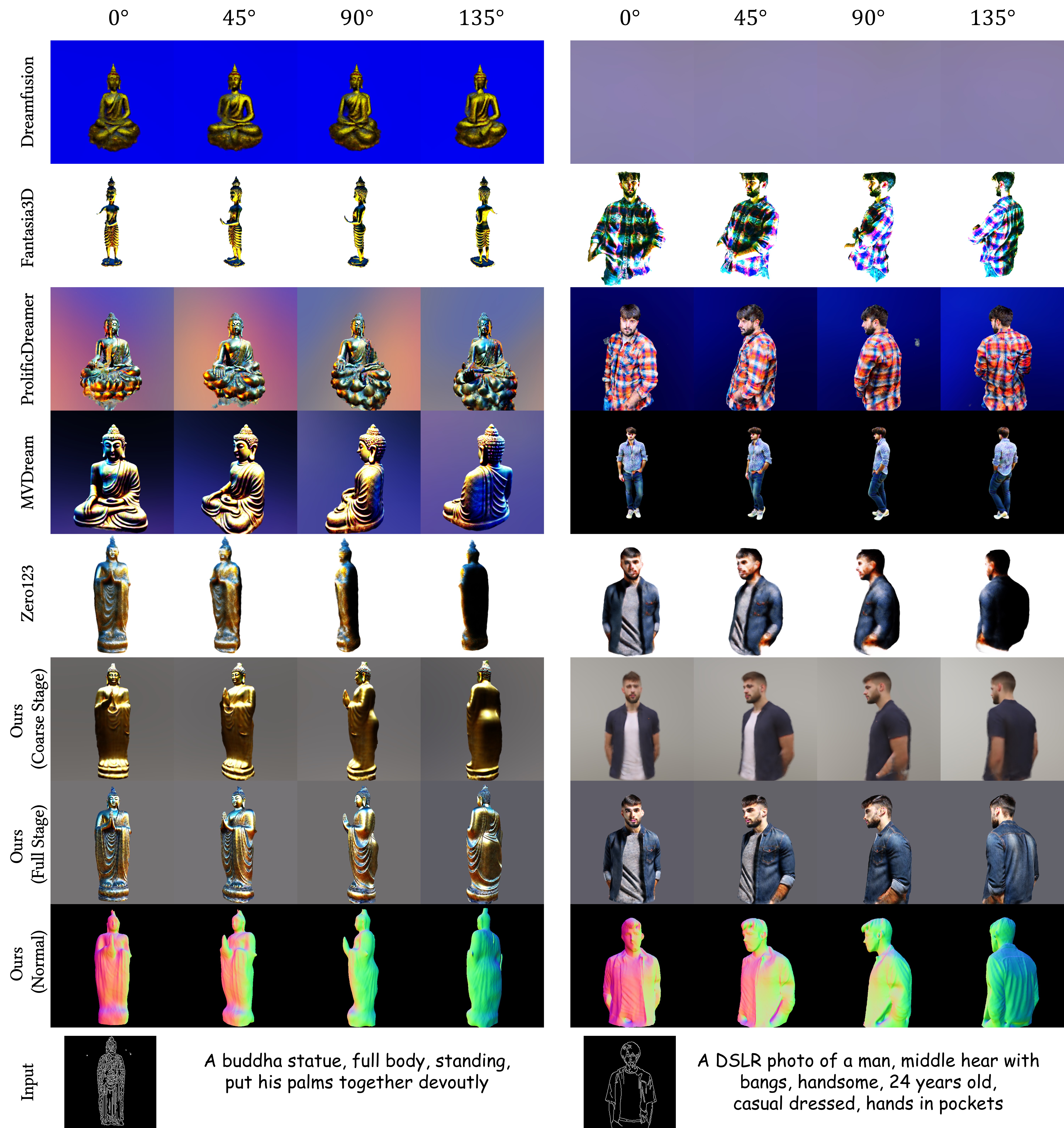}
	\captionof{figure}{\textbf{Controllable text-to-3D content generation.} It demonstrates that our method is able to generate controllable high-fidelity 3D assets over prior methods. We use the rendered reference image of our final model for Zero123 to have proper comparison.}
        \label{fig_comparison}
	\vspace{-0.7em}
\end{figure*}

\PAR{Coarse-to-fine optimization strategy for 3D generation.} We study the benefit to exploit the coarse-to-fine strategy for text-to-3D generation via SDS optimization. The experimental results are shown in \figref{fig_comparison}, \ie \textit{Ours (Coarse Stage)} and \textit{Ours (Full Stage)}. It demonstrates that we can obtain a good geometry of the generated 3D asset during the coarse stage. However, the texture lacks details and is over-smoothed. After we converted the generated 3D asset to deformable mesh \cite{shen2021dmtet} and optimize the texture only at a higher resolution, the 3D asset looks more photo-realistic and contains richer texture details. It demonstrates the necessity of the fine texture optimization stage.


\subsection{Controllable Multi-view Image Generation}\label{subsec:exp_2d}
We compare the performance of our network against its base model, \ie MVDream. The experimental results present in \figref{fig_2d_exp1} demonstrate both networks are able to generate multi-view consistent images, which satisfy the input text prompt description. However, since MVDream cannot accept additional condition image, they are unable to control the shapes of the generated images. In contrary, our method is able to generate controllable multi-view consistent images with Canny edge image as additional input.


\subsection{Controllable 3D Content Generation}\label{subsec:exp_3d}
We compare against prior state-of-the-art text-to-3D generation methods, \ie DreamFusion \cite{poole2022dreamfusion}, Fantasia3D \cite{chen2023fantasia3d}, ProlificDreamer \cite{wang2023prolificdreamer} and MVDream \cite{shi2023mvdream}. The results shown in \figref{fig_comparison} demonstrate that prior works can generate reasonable 3D assets, but suffer from the Janus problem and lack proper control via a condition image. We also compare against an image-based method, \ie Zero123 \cite{liu2023zero123}. For proper comparison, we render the reference image of our generated asset after the fine stage for Zero123 \cite{liu2023zero123}. The results demonstrate that it can generate proper 3D asset satisfying the edge map. However, it lacks details at the back of the 3D assets. We use their default setup implemented in the threestudio repository. It demonstrates that our method is able to generate controllable high-fidelity 3D assets over prior methods.



	\section{Conclusion}\label{sec:con}
We present a novel network architecture for controllable text-to-multiview image generation. Our network exploits a pretrained image diffusion network as base model. A novel trainable control network is proposed to interact with the base model for controllable multiview image generation. Once it is trained, our network can provide 3D prior for controllable text-to-3D generation via SDS optimization. The experimental results demonstrate that our method can generate controllable high-fidelity text-to-multiview images and text-to-3D assets. Besides being used for controllable 3D generation via SDS optimization, we believe our network would be applicable for more broad 3D vision/graphic application scenarios in future.
	{
		\small
		\bibliographystyle{ieee_fullname}
		\bibliography{bibliography_zhiqi,bibliography_liu}

\begin{thebibliography}{10}\itemsep=-1pt

\bibitem{achlioptas2018learning}
Panos Achlioptas, Olga Diamanti, Ioannis Mitliagkas, and Leonidas Guibas.
\newblock Learning representations and generative models for 3d point clouds.
\newblock In {\em International conference on machine learning}, pages 40--49. PMLR, 2018.

\bibitem{pumarola2020cvpr}
Francesc Moreno-Noguer Albert~Pumarola, Stefan~Popov and Vittorio Ferrari.
\newblock {C-Flow: Conditional generative flow models for images and 3D point clouds}.
\newblock In {\em CVPR}, 2020.

\bibitem{aliakbarian2022cvpr}
Sadegh Aliakbarian, Pashmina Cameron, Federica Bogo, Andrew Fitzgibbon, and Thomas~J. Cashman.
\newblock {FLAG: Flow-based 3D Avatar generation from sparse observations}.
\newblock In {\em CVPR}, 2022.

\bibitem{armandpour2023prepneg}
Mohammadreza Armandpour, Huangjie Zheng, Ali Sadeghian, Amir Sadeghian, and Mingyuan Zhou.
\newblock Re-imagine the negative prompt algorithm: Transform 2d diffusion into 3d, alleviate janus problem and beyond.
\newblock {\em arXiv preprint arXiv:2304.04968}, 2023.

\bibitem{balashova2018structure}
Elena Balashova, Vivek Singh, Jiangping Wang, Brian Teixeira, Terrence Chen, and Thomas Funkhouser.
\newblock Structure-aware shape synthesis.
\newblock In {\em 2018 International Conference on 3D Vision (3DV)}, pages 140--149. IEEE, 2018.

\bibitem{ben2018multi}
Heli Ben-Hamu, Haggai Maron, Itay Kezurer, Gal Avineri, and Yaron Lipman.
\newblock Multi-chart generative surface modeling.
\newblock {\em ACM Transactions on Graphics (TOG)}, 37(6):1--15, 2018.

\bibitem{blender}
{Blender Online Community}.
\newblock {\em Blender - a 3D modelling and rendering package}.
\newblock Blender Foundation, Blender Institute, Amsterdam.

\bibitem{brock20163dgan}
Andrew Brock, Theodore Lim, J.M. Ritchie, and Nick Weston.
\newblock {Generative and discriminative voxel modeling with convolutional neural networks}.
\newblock In {\em NeurIPS}, 2016.

\bibitem{chan2022eg3d}
Eric~R Chan, Connor~Z Lin, Matthew~A Chan, Koki Nagano, Boxiao Pan, Shalini De~Mello, Orazio Gallo, Leonidas~J Guibas, Jonathan Tremblay, Sameh Khamis, et~al.
\newblock Efficient geometry-aware 3d generative adversarial networks.
\newblock In {\em Proceedings of the IEEE/CVF Conference on Computer Vision and Pattern Recognition}, pages 16123--16133, 2022.

\bibitem{chan20233diffusion}
Eric~R. Chan, Koki Nagano, Matthew~A. Chan, Alexander~W. Bergman, Jeong~Joon Park, Axel Levy, Miika Aittala, Shalini~De Mello, Tero Karras, and Gordon Wetzstein.
\newblock {Generative novel view synthesis with 3D aware diffusion models}.
\newblock In {\em ICCV}, 2023.

\bibitem{shapenet2015}
Angel~X. Chang, Thomas Funkhouser, Leonidas Guibas, Pat Hanrahan, Qixing Huang, Zimo Li, Silvio Savarese, Manolis Savva, Shuran Song, Hao Su, Jianxiong Xiao, Li Yi, and Fisher Yu.
\newblock {ShapeNet: An Information-Rich 3D Model Repository}.
\newblock Technical Report arXiv:1512.03012 [cs.GR], Stanford University --- Princeton University --- Toyota Technological Institute at Chicago, 2015.

\bibitem{chen2023fantasia3d}
Rui Chen, Yongwei Chen, Ningxin Jiao, and Kui Jia.
\newblock Fantasia3d: Disentangling geometry and appearance for high-quality text-to-3d content creation.
\newblock {\em arXiv preprint arXiv:2303.13873}, 2023.

\bibitem{chen2019cvpr}
Zhiqin Chen and Hao Zhang.
\newblock {Learning implicit fields for generative shape modeling}.
\newblock In {\em CVPR}, 2019.

\bibitem{chou2023diffsdf}
Gene Chou, Yuval Bahat, and Felix Heide.
\newblock {DiffusionSDF: Conditional generative modeling of signed distance functions}.
\newblock In {\em ICCV}, 2023.

\bibitem{pavllo2021iccv}
Thomas~Hofmann Dario~Pavllo, Jonas~Kohler and Aurelien Lucchi.
\newblock {Learning generative models of textured 3D meshes from real-world images}.
\newblock In {\em ICCV}, 2021.

\bibitem{deitke2023objaverse}
Matt Deitke, Dustin Schwenk, Jordi Salvador, Luca Weihs, Oscar Michel, Eli VanderBilt, Ludwig Schmidt, Kiana Ehsani, Aniruddha Kembhavi, and Ali Farhadi.
\newblock Objaverse: A universe of annotated 3d objects.
\newblock In {\em Proceedings of the IEEE/CVF Conference on Computer Vision and Pattern Recognition}, pages 13142--13153, 2023.

\bibitem{gadelha20173d}
Matheus Gadelha, Subhransu Maji, and Rui Wang.
\newblock 3d shape induction from 2d views of multiple objects.
\newblock In {\em 2017 International Conference on 3D Vision (3DV)}, pages 402--411. IEEE, 2017.

\bibitem{gao2019tog}
Lin Gao, Jie Yang, Tong Wu, Yujie Yuan, Hongbo Fu, Yukun Lai, and Hao Zhang.
\newblock {SDM-Net: Deep generative network for structured deformable mesh}.
\newblock In {\em ACM TOG}, 2019.

\bibitem{accelerate}
Sylvain Gugger, Lysandre Debut, Thomas Wolf, Philipp Schmid, Zachary Mueller, Sourab Mangrulkar, Marc Sun, and Benjamin Bossan.
\newblock Accelerate: Training and inference at scale made simple, efficient and adaptable.
\newblock \url{https://github.com/huggingface/accelerate}, 2022.

\bibitem{threestudio2023}
Yuan-Chen Guo, Ying-Tian Liu, Ruizhi Shao, Christian Laforte, Vikram Voleti, Guan Luo, Chia-Hao Chen, Zi-Xin Zou, Chen Wang, Yan-Pei Cao, and Song-Hai Zhang.
\newblock threestudio: A unified framework for 3d content generation.
\newblock \url{https://github.com/threestudio-project/threestudio}, 2023.

\bibitem{ho2020denoising}
Jonathan Ho, Ajay Jain, and Pieter Abbeel.
\newblock Denoising diffusion probabilistic models.
\newblock {\em Advances in neural information processing systems}, 33:6840--6851, 2020.

\bibitem{ho2022cfg}
Jonathan Ho and Tim Salimans.
\newblock Classifier-free diffusion guidance.
\newblock {\em arXiv preprint arXiv:2207.12598}, 2022.

\bibitem{huang2023dreamtime}
Yukun Huang, Jianan Wang, Yukai Shi, Xianbiao Qi, Zheng-Jun Zha, and Lei Zhang.
\newblock Dreamtime: An improved optimization strategy for text-to-3d content creation.
\newblock {\em arXiv preprint arXiv:2306.12422}, 2023.

\bibitem{hui2022neural}
Ka-Hei Hui, Ruihui Li, Jingyu Hu, and Chi-Wing Fu.
\newblock Neural wavelet-domain diffusion for 3d shape generation.
\newblock In {\em SIGGRAPH Asia 2022 Conference Papers}, pages 1--9, 2022.

\bibitem{jain2022zero}
Ajay Jain, Ben Mildenhall, Jonathan~T Barron, Pieter Abbeel, and Ben Poole.
\newblock Zero-shot text-guided object generation with dream fields.
\newblock In {\em Proceedings of the IEEE/CVF Conference on Computer Vision and Pattern Recognition}, pages 867--876, 2022.

\bibitem{katzir2023nfsd}
Oren Katzir, Or Patashnik, Daniel Cohen-Or, and Dani Lischinski.
\newblock Noise-free score distillation.
\newblock {\em arXiv preprint arXiv:2310.17590}, 2023.

\bibitem{kerbl20233dgaussian}
Bernhard Kerbl, Georgios Kopanas, Thomas Leimk{\"u}hler, and George Drettakis.
\newblock 3d gaussian splatting for real-time radiance field rendering.
\newblock {\em ACM Transactions on Graphics (ToG)}, 42(4):1--14, 2023.

\bibitem{kingma2014adam}
Diederik~P Kingma and Jimmy Ba.
\newblock Adam: A method for stochastic optimization.
\newblock {\em arXiv preprint arXiv:1412.6980}, 2014.

\bibitem{klokov2020eccv}
Roman Klokov, Edmond Boyer, and Jakob Verbeek.
\newblock {Discrete point flow networks for efficient point cloud generation}.
\newblock In {\em ECCV}, 2020.

\bibitem{li2023sweetdreamer}
Weiyu Li, Rui Chen, Xuelin Chen, and Ping Tan.
\newblock Sweetdreamer: Aligning geometric priors in 2d diffusion for consistent text-to-3d.
\newblock {\em arXiv preprint arXiv:2310.02596}, 2023.

\bibitem{lin2023magic3d}
Chen-Hsuan Lin, Jun Gao, Luming Tang, Towaki Takikawa, Xiaohui Zeng, Xun Huang, Karsten Kreis, Sanja Fidler, Ming-Yu Liu, and Tsung-Yi Lin.
\newblock Magic3d: High-resolution text-to-3d content creation.
\newblock In {\em Proceedings of the IEEE/CVF Conference on Computer Vision and Pattern Recognition}, pages 300--309, 2023.

\bibitem{lin2023common}
Shanchuan Lin, Bingchen Liu, Jiashi Li, and Xiao Yang.
\newblock Common diffusion noise schedules and sample steps are flawed.
\newblock {\em arXiv preprint arXiv:2305.08891}, 2023.

\bibitem{liu2023zero123}
Ruoshi Liu, Rundi Wu, Basile Van~Hoorick, Pavel Tokmakov, Sergey Zakharov, and Carl Vondrick.
\newblock Zero-1-to-3: Zero-shot one image to 3d object.
\newblock In {\em Proceedings of the IEEE/CVF International Conference on Computer Vision}, pages 9298--9309, 2023.

\bibitem{liu2023syncdreamer}
Yuan Liu, Cheng Lin, Zijiao Zeng, Xiaoxiao Long, Lingjie Liu, Taku Komura, and Wenping Wang.
\newblock Syncdreamer: Generating multiview-consistent images from a single-view image.
\newblock {\em arXiv preprint arXiv:2309.03453}, 2023.

\bibitem{long2023wonder3d}
Xiaoxiao Long, Yuan-Chen Guo, Cheng Lin, Yuan Liu, Zhiyang Dou, Lingjie Liu, Yuexin Ma, Song-Hai Zhang, Marc Habermann, Christian Theobalt, et~al.
\newblock Wonder3d: Single image to 3d using cross-domain diffusion.
\newblock {\em arXiv preprint arXiv:2310.15008}, 2023.

\bibitem{luo2021cvpr}
Shitong Luo and Wei Hu.
\newblock {Diffusion Probabilistic Models for 3D Point Cloud Generation}.
\newblock In {\em CVPR}, 2021.

\bibitem{luo2023cap3d}
Tiange Luo, Chris Rockwell, Honglak Lee, and Justin Johnson.
\newblock Scalable 3d captioning with pretrained models.
\newblock {\em arXiv preprint arXiv:2306.07279}, 2023.

\bibitem{mescheder2019cvpr}
Lars Mescheder, Michael Oechsle, Michael Niemeyer, Sebastuan Nowozin, and Andreas Geiger.
\newblock {Occupancy Networks: Learning 3D reconstruction in function space}.
\newblock In {\em CVPR}, 2019.

\bibitem{metzer2023latent}
Gal Metzer, Elad Richardson, Or Patashnik, Raja Giryes, and Daniel Cohen-Or.
\newblock Latent-nerf for shape-guided generation of 3d shapes and textures.
\newblock In {\em Proceedings of the IEEE/CVF Conference on Computer Vision and Pattern Recognition}, pages 12663--12673, 2023.

\bibitem{mildenhall2021nerf}
Ben Mildenhall, Pratul~P Srinivasan, Matthew Tancik, Jonathan~T Barron, Ravi Ramamoorthi, and Ren Ng.
\newblock Nerf: Representing scenes as neural radiance fields for view synthesis.
\newblock {\em Communications of the ACM}, 65(1):99--106, 2021.

\bibitem{mohammad2022clip}
Nasir Mohammad~Khalid, Tianhao Xie, Eugene Belilovsky, and Tiberiu Popa.
\newblock Clip-mesh: Generating textured meshes from text using pretrained image-text models.
\newblock In {\em SIGGRAPH Asia 2022 conference papers}, pages 1--8, 2022.

\bibitem{muller2023diffrf}
Norman Muller, Yawar Siddiqui, Lorenzo Porzi, Samuel~Rota Bulo, Peter Kontschieder, and Matthias Niebner.
\newblock {DiffRF: Rendering-Guided 3D Radiance Field Diffusion}.
\newblock In {\em CVPR}, 2023.

\bibitem{muller2022instant}
Thomas M{\"u}ller, Alex Evans, Christoph Schied, and Alexander Keller.
\newblock Instant neural graphics primitives with a multiresolution hash encoding.
\newblock {\em ACM Transactions on Graphics (ToG)}, 41(4):1--15, 2022.

\bibitem{phuoc2020neurips}
Thu~H Nguyen-Phuoc, Christian Richardt, Long Mai, Yongliang Yang, and Niloy Mitra.
\newblock Blockgan: Learning 3d object-aware scene representations from unlabelled images.
\newblock {\em Advances in neural information processing systems}, 33:6767--6778, 2020.

\bibitem{nichol2022point}
Alex Nichol, Heewoo Jun, Prafulla Dhariwal, Pamela Mishkin, and Mark Chen.
\newblock Point-e: A system for generating 3d point clouds from complex prompts.
\newblock {\em arXiv preprint arXiv:2212.08751}, 2022.

\bibitem{park2019cvpr}
Jeong~Joon Park, Peter Florence, Julian Straub, Richard Newcombe, and Steven Lovegrove.
\newblock {DeepSDF: Learning continuous signed distance functions for shape representation}.
\newblock In {\em CVPR}, 2019.

\bibitem{paszke2019pytorch}
Adam Paszke, Sam Gross, Francisco Massa, Adam Lerer, James Bradbury, Gregory Chanan, Trevor Killeen, Zeming Lin, Natalia Gimelshein, Luca Antiga, et~al.
\newblock Pytorch: An imperative style, high-performance deep learning library.
\newblock {\em Advances in neural information processing systems}, 32, 2019.

\bibitem{poole2022dreamfusion}
Ben Poole, Ajay Jain, Jonathan~T Barron, and Ben Mildenhall.
\newblock Dreamfusion: Text-to-3d using 2d diffusion.
\newblock In {\em ICLR}, 2023.

\bibitem{radford2021learning}
Alec Radford, Jong~Wook Kim, Chris Hallacy, Aditya Ramesh, Gabriel Goh, Sandhini Agarwal, Girish Sastry, Amanda Askell, Pamela Mishkin, Jack Clark, et~al.
\newblock Learning transferable visual models from natural language supervision.
\newblock In {\em International conference on machine learning}, pages 8748--8763. PMLR, 2021.

\bibitem{raj2023dreambooth3d}
Amit Raj, Srinivas Kaza, Ben Poole, Michael Niemeyer, Ben Mildenhall, Nataniel Ruiz, Shiran Zada, Kfir Aberman, Michael Rubenstein, Jonathan Barron, Yuanzhen Li, and Varun Jampani.
\newblock Dreambooth3d: Subject-driven text-to-3d generation.
\newblock In {\em ICCV}, 2023.

\bibitem{ranftl2021vision}
Ren{\'e} Ranftl, Alexey Bochkovskiy, and Vladlen Koltun.
\newblock Vision transformers for dense prediction.
\newblock In {\em Proceedings of the IEEE/CVF international conference on computer vision}, pages 12179--12188, 2021.

\bibitem{rombach2022stablediffusion}
Robin Rombach, Andreas Blattmann, Dominik Lorenz, Patrick Esser, and Bj{\"o}rn Ommer.
\newblock High-resolution image synthesis with latent diffusion models.
\newblock In {\em Proceedings of the IEEE/CVF conference on computer vision and pattern recognition}, pages 10684--10695, 2022.

\bibitem{saharia2022photorealistic}
Chitwan Saharia, William Chan, Saurabh Saxena, Lala Li, Jay Whang, Emily~L Denton, Kamyar Ghasemipour, Raphael Gontijo~Lopes, Burcu Karagol~Ayan, Tim Salimans, et~al.
\newblock Photorealistic text-to-image diffusion models with deep language understanding.
\newblock {\em Advances in Neural Information Processing Systems}, 35:36479--36494, 2022.

\bibitem{sanghi2022clipforge}
Aditya Sanghi, Hang Chu, Joseph~G. Lambourne, Ye Wang, Chinyi Cheng, Marco Fumero, and Kamal~Rahimi Malekshan.
\newblock {CLIP-Forge: Towards Zero-Shot Text-to-Shape Generation}.
\newblock In {\em Proceedings of the IEEE/CVF Conference on Computer Vision and Pattern Recognition}, 2022.

\bibitem{schuhmann2022laion}
Christoph Schuhmann, Romain Beaumont, Richard Vencu, Cade Gordon, Ross Wightman, Mehdi Cherti, Theo Coombes, Aarush Katta, Clayton Mullis, Mitchell Wortsman, et~al.
\newblock Laion-5b: An open large-scale dataset for training next generation image-text models.
\newblock {\em Advances in Neural Information Processing Systems}, 35:25278--25294, 2022.

\bibitem{schwarz2022neurips}
Katja Schwarz, Axel Sauer, Michael Niemeyer, Yiyi Liao, and Andreas Geiger.
\newblock {VoxGRAF: Fast 3D-aware image synthesis with sparse voxel grids}.
\newblock In {\em NeurIPS}, 2022.

\bibitem{shen2021dmtet}
Tianchang Shen, Jun Gao, Kangxue Yin, Ming-Yu Liu, and Sanja Fidler.
\newblock Deep marching tetrahedra: a hybrid representation for high-resolution 3d shape synthesis.
\newblock {\em Advances in Neural Information Processing Systems}, 34:6087--6101, 2021.

\bibitem{shi2023zero123++}
Ruoxi Shi, Hansheng Chen, Zhuoyang Zhang, Minghua Liu, Chao Xu, Xinyue Wei, Linghao Chen, Chong Zeng, and Hao Su.
\newblock Zero123++: a single image to consistent multi-view diffusion base model.
\newblock {\em arXiv preprint arXiv:2310.15110}, 2023.

\bibitem{shi2023mvdream}
Yichun Shi, Peng Wang, Jianglong Ye, Mai Long, Kejie Li, and Xiao Yang.
\newblock Mvdream: Multi-view diffusion for 3d generation.
\newblock {\em arXiv preprint arXiv:2308.16512}, 2023.

\bibitem{shu20193d}
Dong~Wook Shu, Sung~Woo Park, and Junseok Kwon.
\newblock 3d point cloud generative adversarial network based on tree structured graph convolutions.
\newblock In {\em Proceedings of the IEEE/CVF international conference on computer vision}, pages 3859--3868, 2019.

\bibitem{song2020score}
Yang Song, Jascha Sohl-Dickstein, Diederik~P Kingma, Abhishek Kumar, Stefano Ermon, and Ben Poole.
\newblock Score-based generative modeling through stochastic differential equations.
\newblock {\em arXiv preprint arXiv:2011.13456}, 2020.

\bibitem{sun2023dreamcraft3d}
Jingxiang Sun, Bo Zhang, Ruizhi Shao, Lizhen Wang, Wen Liu, Zhenda Xie, and Yebin Liu.
\newblock Dreamcraft3d: Hierarchical 3d generation with bootstrapped diffusion prior.
\newblock {\em arXiv preprint arXiv:2310.16818}, 2023.

\bibitem{sun2023unig3d}
Qinghong Sun, Yangguang Li, ZeXiang Liu, Xiaoshui Huang, Fenggang Liu, Xihui Liu, Wanli Ouyang, and Jing Shao.
\newblock Unig3d: A unified 3d object generation dataset.
\newblock {\em arXiv preprint arXiv:2306.10730}, 2023.

\bibitem{tan2018cvpr}
Qingyang Tan, Lin Gao, Yukun Lai, and Shihong Xia.
\newblock {Variational autoencoders for deforming 3D mesh models}.
\newblock In {\em CVPR}, 2018.

\bibitem{tang2023dreamgaussian}
Jiaxiang Tang, Jiawei Ren, Hang Zhou, Ziwei Liu, and Gang Zeng.
\newblock Dreamgaussian: Generative gaussian splatting for efficient 3d content creation.
\newblock {\em arXiv preprint arXiv:2309.16653}, 2023.

\bibitem{tang2023make}
Junshu Tang, Tengfei Wang, Bo Zhang, Ting Zhang, Ran Yi, Lizhuang Ma, and Dong Chen.
\newblock Make-it-3d: High-fidelity 3d creation from a single image with diffusion prior.
\newblock {\em arXiv preprint arXiv:2303.14184}, 2023.

\bibitem{tang2023mvdiffusion}
Shitao Tang, Fuyang Zhang, Jiacheng Chen, Peng Wang, and Yasutaka Furukawa.
\newblock {MVDiffusion: enabling holistic multi-view image generation with correspondence aware diffusion}.
\newblock In {\em NeurIPS}, 2023.

\bibitem{tseng2023novelview}
Hung-Yu Tseng, Qinbo Li, Changil Kim, Suhib Alsisan, Jiabin Huang, and Johannes Kopf.
\newblock {Consistent view synthesis with pose guided diffusion models}.
\newblock In {\em CVPR}, 2023.

\bibitem{diffusers}
Patrick von Platen, Suraj Patil, Anton Lozhkov, Pedro Cuenca, Nathan Lambert, Kashif Rasul, Mishig Davaadorj, and Thomas Wolf.
\newblock Diffusers: State-of-the-art diffusion models.
\newblock \url{https://github.com/huggingface/diffusers}, 2022.

\bibitem{wang2023SJC}
Haochen Wang, Xiaodan Du, Jiahao Li, Raymond~A. Yeh, and Greg Shakhnarovich.
\newblock {Score Jacobian Chaining: lifting pretrained 2D diffusion models for 3D generation}.
\newblock In {\em Proceedings of the IEEE/CVF Conference on Computer Vision and Pattern Recognition}, 2023.

\bibitem{wang2021neus}
Peng Wang, Lingjie Liu, Yuan Liu, Christian Theobalt, Taku Komura, and Wenping Wang.
\newblock Neus: Learning neural implicit surfaces by volume rendering for multi-view reconstruction.
\newblock {\em arXiv preprint arXiv:2106.10689}, 2021.

\bibitem{wang2023rodin}
Tengfei Wang, Bo Zhang, Ting Zhang, Shuyang Gu, Jianmin Bao, Tadas Baltrusaitis, Jingjing Shen, Dong Chen, Fang Wen, Qifeng Chen, et~al.
\newblock Rodin: A generative model for sculpting 3d digital avatars using diffusion.
\newblock In {\em Proceedings of the IEEE/CVF Conference on Computer Vision and Pattern Recognition}, pages 4563--4573, 2023.

\bibitem{wang2023prolificdreamer}
Zhengyi Wang, Cheng Lu, Yikai Wang, Fan Bao, Chongxuan Li, Hang Su, and Jun Zhu.
\newblock Prolificdreamer: High-fidelity and diverse text-to-3d generation with variational score distillation.
\newblock {\em arXiv preprint arXiv:2305.16213}, 2023.

\bibitem{watson2023novelview}
Daniel Watson, William Chan, Ricardo Martin-Brualla, Jonathan Ho, Andrea Tagliasacchi, and Mohammad Norouzi.
\newblock {Novel View Synthesis with Diffusion Models}.
\newblock In {\em ICLR}, 2023.

\bibitem{transformers}
Thomas Wolf, Lysandre Debut, Victor Sanh, Julien Chaumond, Clement Delangue, Anthony Moi, Pierric Cistac, Tim Rault, Rémi Louf, Morgan Funtowicz, Joe Davison, Sam Shleifer, Patrick von Platen, Clara Ma, Yacine Jernite, Julien Plu, Canwen Xu, Teven~Le Scao, Sylvain Gugger, Mariama Drame, Quentin Lhoest, and Alexander~M. Rush.
\newblock Transformers: State-of-the-art natural language processing.
\newblock In {\em Proceedings of the 2020 Conference on Empirical Methods in Natural Language Processing: System Demonstrations}, pages 38--45, Online, Oct. 2020. Association for Computational Linguistics.

\bibitem{wu2016learning}
Jiajun Wu, Chengkai Zhang, Tianfan Xue, Bill Freeman, and Josh Tenenbaum.
\newblock Learning a probabilistic latent space of object shapes via 3d generative-adversarial modeling.
\newblock {\em Advances in neural information processing systems}, 29, 2016.

\bibitem{wu2019ToG}
Zhijie Wu, Xiang Wang, Di Lin, Dani Lischinski, Daniel Cohen-Or, and Hui Huang.
\newblock {SAGNet: Structure aware generative network for 3D shape modeling}.
\newblock In {\em ACM TOG}, 2019.

\bibitem{xiang20233daware}
Jianfeng Xiang, Jiaolong Yang, Binbin Huang, and Xin Tong.
\newblock {3D-aware image generation using 2D diffusion models}.
\newblock In {\em ICCV}, 2023.

\bibitem{li2019gan}
Pieter~Peers Xiao~Li, Yue~Dong and Xin Tong.
\newblock {Synthesizing 3D shapes from silhouette image collections using multi-projection generative adversarial networks}.
\newblock In {\em CVPR}, 2019.

\bibitem{yang2019iccv}
Guandao Yang, Xun Huang, Zekun Hao, Mingyu Liu, Serge Belongie, and Bharath Hariharan.
\newblock {PointFlow: 3D Point Cloud Generation with Continuous Normalizing Flows}.
\newblock In {\em ICCV}, 2019.

\bibitem{yang2019pointflow}
Guandao Yang, Xun Huang, Zekun Hao, Ming-Yu Liu, Serge Belongie, and Bharath Hariharan.
\newblock Pointflow: 3d point cloud generation with continuous normalizing flows.
\newblock In {\em Proceedings of the IEEE/CVF international conference on computer vision}, pages 4541--4550, 2019.

\bibitem{youwang2022eccv}
Kim Youwang, Kim Ji-Yeon, and Tae-Hyun Oh.
\newblock {CLIP-Actor: text driven recommendation and stylization for animating human meshes}.
\newblock In {\em ECCV}, 2022.

\bibitem{yu2023csd}
Xin Yu, Yuan-Chen Guo, Yangguang Li, Ding Liang, Song-Hai Zhang, and Xiaojuan Qi.
\newblock Text-to-3d with classifier score distillation.
\newblock {\em arXiv preprint arXiv:2310.19415}, 2023.

\bibitem{zeng2022neurips}
Xiaohui Zeng, Arash Vahdat, Francis Williams, Zan Gojcic, Or Litany, Sanja Fidler, and Karsten Kreis.
\newblock {LION: latent point diffusion models for 3D shape generation}.
\newblock In {\em NeurIPS}, 2022.

\bibitem{zhang2023controlnet}
Lvmin Zhang, Anyi Rao, and Maneesh Agrawala.
\newblock Adding conditional control to text-to-image diffusion models.
\newblock In {\em Proceedings of the IEEE/CVF International Conference on Computer Vision}, pages 3836--3847, 2023.

\bibitem{zhou2021iccv}
Linqi Zhou, Yilun Du, and Jiajun Wu.
\newblock {3D Shape Generation and Completion through Point-Voxel Diffusion}.
\newblock In {\em ICCV}, 2021.

\bibitem{zhu2023hifa}
Joseph Zhu and Peiye Zhuang.
\newblock Hifa: High-fidelity text-to-3d with advanced diffusion guidance.
\newblock {\em arXiv preprint arXiv:2305.18766}, 2023.

\end{thebibliography}
	}

        \newpage

\twocolumn[
\centering
\Large
\textbf{Appendix} \\
\vspace{+1em}
] 

\appendix

\section{Introduction}
\label{sec:sup_intro}
In this supplementary material, we provide more details on the method of controlled text-to-3D generation, and implementations of our work and the compared baselines.
Furthermore, we also train a new MVControl network which is conditioned on the dense depth map, for depth controlled text-to-multiview image generation. Additional qualitative results on both the depth map controlled text-to-multiview image generation, and edge map controlled text-to-3D generation are presented.

\section{Controlled Text-to-3D Generation}

Controllable 3D content generation is realized through a coarse-to-fine optimization process. In particular, we first optimize a coarse neural surface \cite{wang2021neus}, and then conduct texture refinement in the fine optimization stage. The coarse geometry is transformed to a deformable mesh \cite{shen2021dmtet} for the texture refinement at the fine stage. As for the hybrid diffusion prior, the pretrained MVControl network works as a strong consistent geometry guidance at four canonical views of the 3D object, and Stable-Diffusion network provides fine geometry and texture sculpting at the other randomly sampled views.
In the following part, we denote the four canonical views with $\mathcal{V}_*$ and the images rendered under those views as $\mathcal{X}_*\in\mathbb{R}^{4\times H\times W\times C}$.

\subsection{Coarse Geometry Stage}
\begin{figure*}[!htbp]
	\centering
	\includegraphics[width=0.93\linewidth]{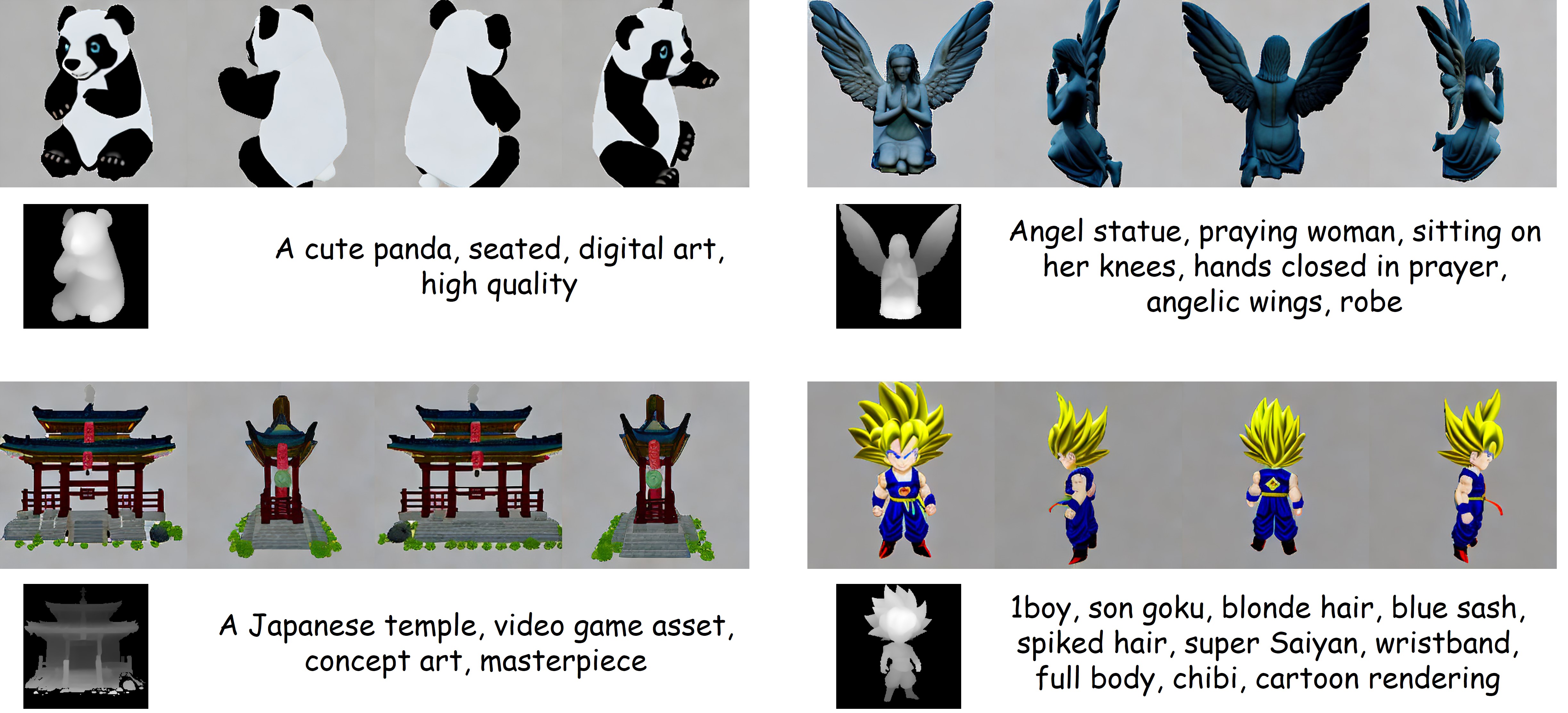}
	\vspace{-0.5em}
	\captionof{figure}{Additional 2D results of depth-conditioned MVControl.}
        \label{fig_supp_2d}
 \vspace{-1em}
\end{figure*}
At this stage, we aim to generate a 3D model whose geometry is consistent with the input condition image. 
While our MVControl can already provide consistent geometry constrains from four canonical views, it's still not sufficient to recover a plausible geometry from the four sparse views only. Hence, we propose to incorporate the 2D diffusion model SD to provide a semantic guidance under those views other than the four canonical views, and so as to sculpt the geometry to satisfy the distribution described by the condition image.

Specifically, we denote a differentiable renderer with $g(\cdot)$ and the parameters of 3D representation as $\theta$. We render the images $\mathcal{X}_*=g(\theta, \mathcal{V}_*)$ under four canonical views and image $x_r=g(\theta, v_r)$ under a randomly sampled view $v_r$. 
Then the gradient of hybrid SDS loss can be computed as:
\begin{equation}
	\nabla_\theta\mathcal{L}_{SDS}^{hybrid}=\lambda_1 \nabla_\theta\mathcal{L}_{SDS}^{2D} + \lambda_2 \nabla_\theta\mathcal{L}_{SDS}^{3D},
\end{equation}
where $\nabla_\theta\mathcal{L}_{SDS}^{2D}$ is the SDS gradient distillated from SD taking the rendering $x_r$ as input, and $\nabla_\theta\mathcal{L}_{SDS}^{3D}$ is that distillated from MVControl network with $\mathcal{X}_*$ and $\mathcal{V}_*$ as input. $\lambda_1$ and $\lambda_2$ are two hyperparameters and chosen empirically. 

While classifier-free guidance (CFG) has become a necessary technique when doing diffusion sampling, we should consider the CFG scale for each of our diffusion priors. 
In order to enforce the optimization process to align with the distribution defined by MVControl, we apply a large CFG scale for $\nabla_\theta\mathcal{L}_{SDS}^{3D}$. 
To avoid the discrepancy between the guidance from both SD (2D) and MVControl (3D), we choose to use a relatively small CFG scale for $\nabla_\theta\mathcal{L}_{SDS}^{2D}$.
Here, following MVDream, we compute $\nabla_\theta\mathcal{L}_{SDS}^{MV}$ through $x_0$-reconstruction formulation to alleviate the color saturation from large CFG scale by applying CFG rescale trick \cite{lin2023common}:
\begin{equation}
	\nabla_\theta\mathcal{L}_{SDS}^{3D}(\psi, \mathcal{X}_*, \mathcal{V}_*)=\mathbb{E}_{t, \epsilon}\lbrack \Vert \mathcal{X}_*-\hat{\mathcal{X}}_0 \Vert _2^2 \rbrack,
\end{equation}
where $\hat{\mathcal{X}}_0$ is the estimated clean images of the four noisy input from $\epsilon_\psi(z_t(\mathcal{X}_*);y, t, \mathcal{V}_*)$ and its gradient is detached from the optimization step. Regarding the computation of $\nabla_\theta\mathcal{L}_{SDS}^{2D}$, we refer to the normal SDS calculation since it uses a small CFG scale. We also exploit the Eikonal loss proposed by \cite{wang2021neus} to regularize the SDF values to be more plausible. 

\subsection{Fine Texture Stage}
In this stage, the coarse geometry is converted to a deformable mesh \cite{shen2021dmtet} for further texture refinement under high rendering resolution with geometry fixed.
For the computation of score distillation gradients, we refer to the recently released Noise-free Score Distillation (NFSD) technique \cite{katzir2023nfsd}. The only difference between our implementation and theirs is that we replace the null prompt $\oslash$ with negative prompt $p_{neg}$ in the $\delta_C$ part, with which we observe a quality improvement. For more details, please refer to \cite{katzir2023nfsd}. Empirically, our method achieves similar results with simple SDS gradient which usually requires large CFG scale, and we choose this strategy due to its normal CFG scales.

\section{Additional Implementation Details}
For training data creation, we use Blender \cite{blender} to render images from Objaverse objects. The rendering scripts are based on a public repository\footnote{\url{https://github.com/allenai/objaverse-rendering}}. The implementations of our model and training code are based on Pytorch \cite{paszke2019pytorch} and heavily rely on the public projects \cite{diffusers, transformers, accelerate} by Hugging Face Organization. 
Our MVControl networks conditioned on edge map and depth map respectively are fine-tuned from public ControlNet checkpoints\footnote{\url{https://huggingface.co/thibaud/controlnet-sd21-canny-diffusers}}\textsuperscript{,}\footnote{\url{https://huggingface.co/thibaud/controlnet-sd21-depth-diffusers}}. And for depth prediction on the training images, we use off-the-shelf depth estimation network \cite{ranftl2021vision}.
The implementations of our 3D generation part and all compared 3D generation baselines are based on the ThreeStudio project \cite{threestudio2023} except MVDream, which is from their official implementation\footnote{\url{https://github.com/bytedance/MVDream-threestudio}}. All the experiments of baselines are conducted under their default setup.    

\section{Additional Qualitative Results}

While we mainly focus on using canny edge maps as the additional condition for MVControl, we also trained another depth-conditioned version of MVControl under the same training settings. Its multi-view image generation results are shown in \figref{fig_supp_2d}. The figure shows that our method is also able to generate high-fidelity multi-view images with depth map as the additional conditioning input together with the text prompt, which demonstrates that our MVControl has the potential to be generalized to different types of conditions.

\begin{figure*}
	\centering
	\includegraphics[width=0.93\linewidth]{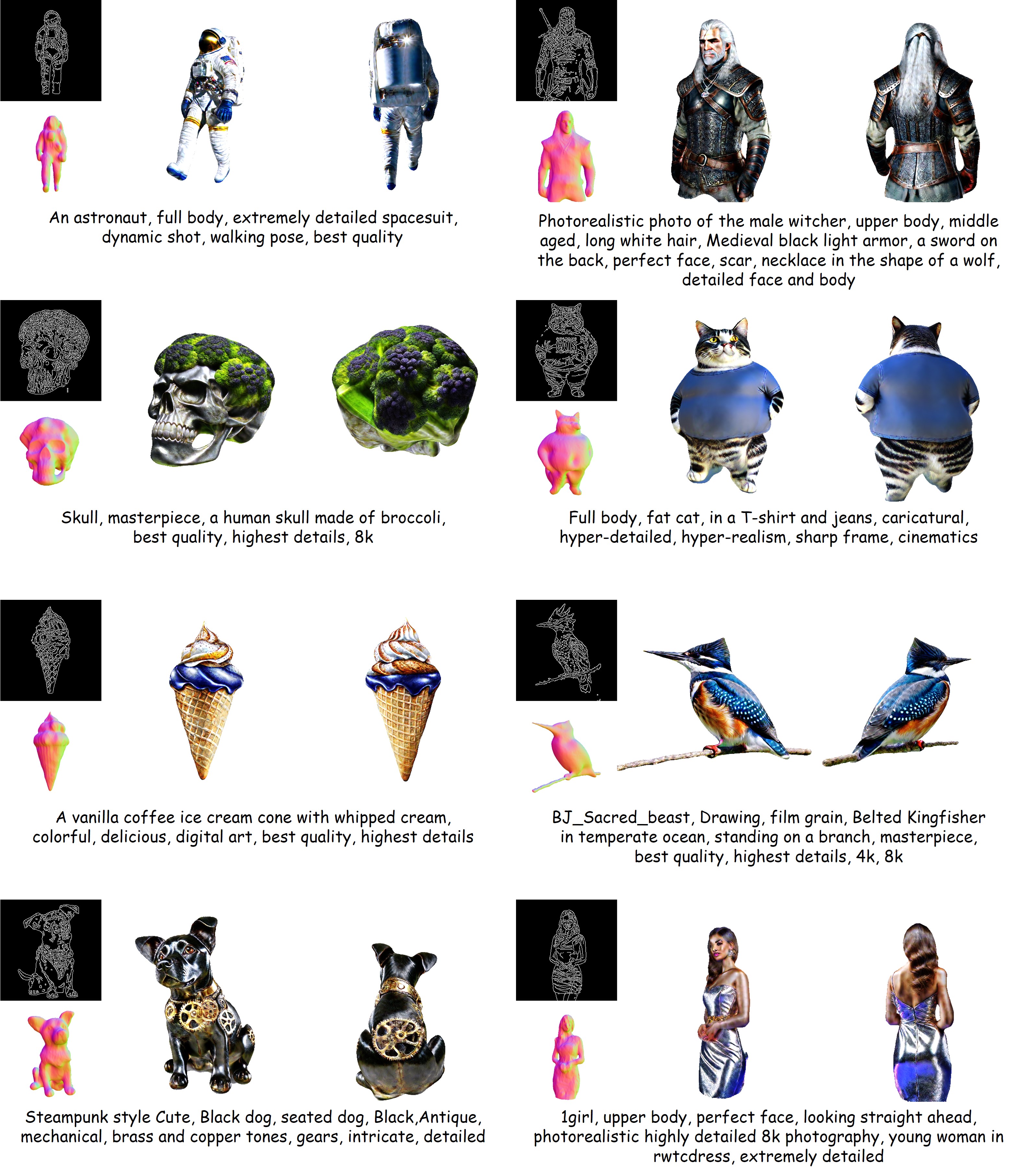}
	\vspace{-0.5em}
	\captionof{figure}{Additional qualitative results of MVControl 3D generation.}
        \label{fig_supp_3d_1}
 \vspace{-1em}
\end{figure*}

\begin{figure*}
	\centering
	\includegraphics[width=0.93\linewidth]{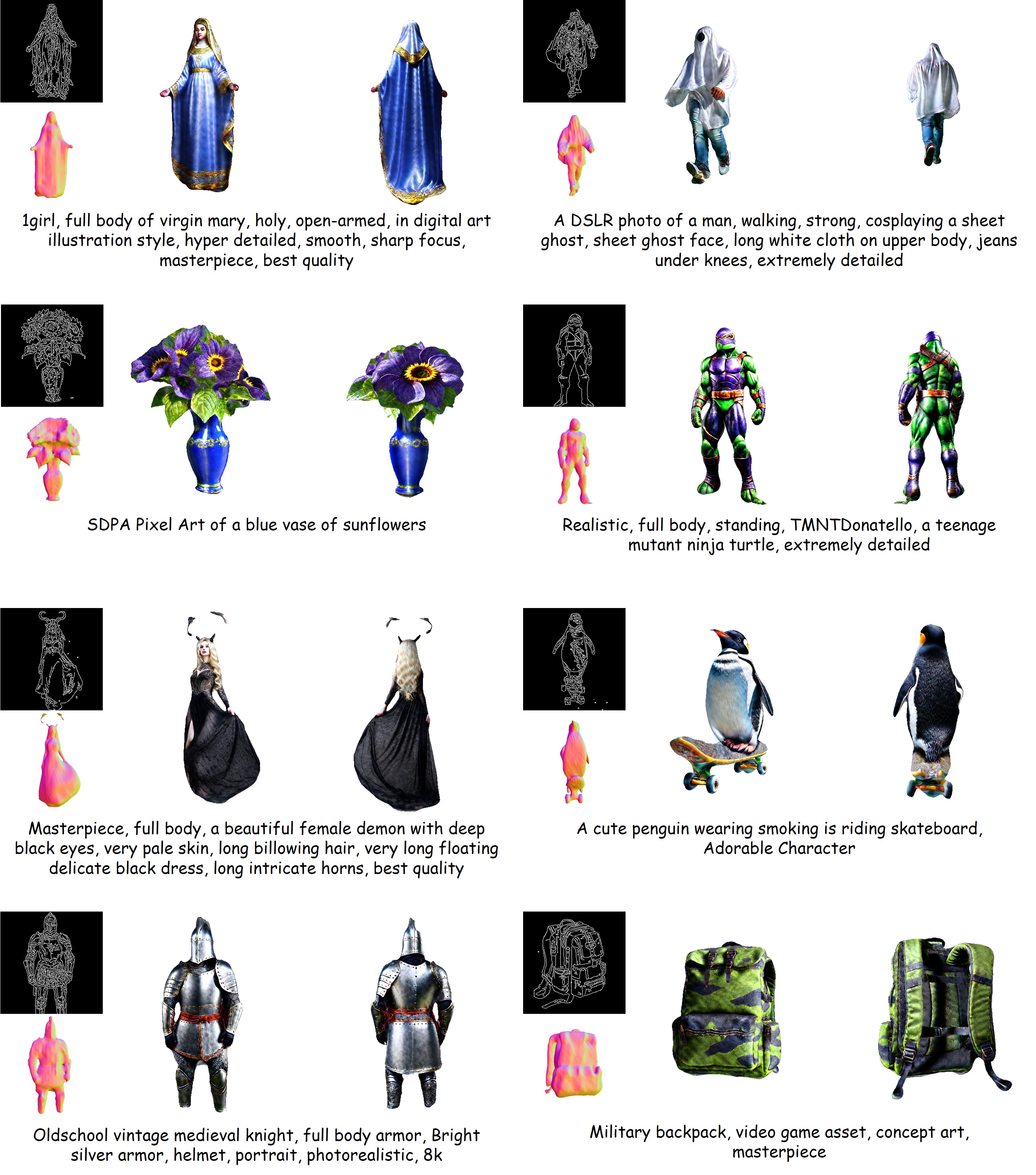}
	\vspace{-0.5em}
	\captionof{figure}{Additional qualitative results of MVControl 3D generation.}
        \label{fig_supp_3d_2}
 \vspace{-1em}
\end{figure*}

We also provide more qualitative results of controlled text-to-3D generation via MVControl in \figref{fig_supp_3d_1} and \figref{fig_supp_3d_2}. The results demonstrate that our method has the capacity to generate high-fidelity view-consistent 3D assets with high-quality texture, which can be controlled by both the text prompt and additional control input (\eg edge map). The readers can refer to our \href{https://lizhiqi49.github.io/MVControl}{project page} for more results.

\end{document}